\documentclass[letterpaper, 10 pt, conference]{ieeeconf}  % Comment this line out if you need a4paper

\IEEEoverridecommandlockouts                              % This command is only needed if 
\usepackage{graphics} % for pdf, bitmapped graphics files
\usepackage{mathptmx} % assumes new font selection scheme installed
\usepackage{times} % assumes new font selection scheme installed
\usepackage{xcolor}
\usepackage{graphicx}
\usepackage[ruled]{algorithm2e}
\usepackage{times}
\usepackage{epsfig}
\usepackage{amsmath}
\usepackage{amssymb}
\usepackage{color}
\usepackage{wrapfig,lipsum,booktabs}
\usepackage{subfig}

\definecolor{citecolor}{RGB}{51, 141, 150}
\usepackage[colorlinks=true,linkcolor=blue, citecolor=citecolor]{hyperref}%
\usepackage{cite}
\usepackage{balance}

\newlength\savewidth\newcommand\shline{\noalign{\global\savewidth\arrayrulewidth
  \global\arrayrulewidth 1pt}\hline\noalign{\global\arrayrulewidth\savewidth}}
\newcommand{\tablestyle}[2]{\setlength{\tabcolsep}{#1}\renewcommand{\arraystretch}{#2}\centering\footnotesize}

\newcommand\blfootnote[1]{%
  \begingroup
  \renewcommand\thefootnote{}\footnote{#1}%
  \addtocounter{footnote}{-1}%
  \endgroup
}

\newcommand{\etal}{\textit{et al}. }

\title{\LARGE \bf
From One Hand to Multiple Hands: Imitation Learning for Dexterous Manipulation from Single-Camera Teleoperation
}

\author{
Yuzhe Qin, \quad
Hao Su\textsuperscript{$\dagger$}, \quad
Xiaolong Wang\textsuperscript{$\dagger$}\\
UC San Diego}

\begin{document}

\twocolumn[{%
\renewcommand\twocolumn[1][]{#1}%
\maketitle
\begin{center}
    \vspace{-0.1in}
    \centering
    \captionsetup{type=figure}
    \includegraphics[width=0.99\linewidth]{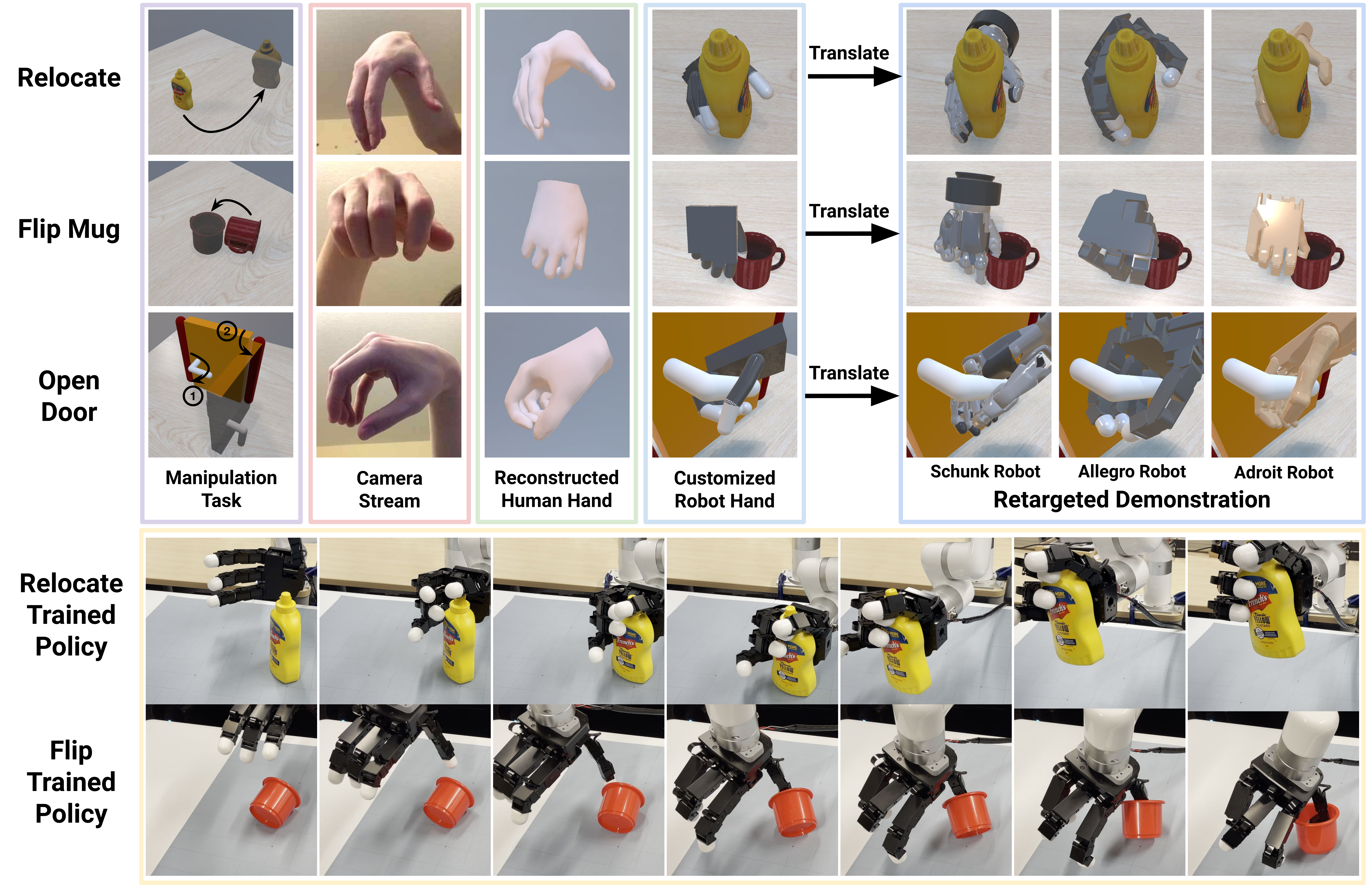}
    \vspace{-0.05in}
    \captionof{figure}{{\small{\textbf{Overview:} We introduce a teleoperation system which utilizes a single camera on an iPad to stream a human hand, estimates the hand pose and shape, and converts it to a customized robot hand in a physical simulator for dexterous manipulation. Once the manipulation trajectories are collected,  we translate them to different specified robot hands to generate demonstrations, and use them to perform imitation learning on the same manipulation tasks. Once the policy is trained, we deploy it to the real robot hand and show robust transfer results.}}}
    \label{fig:teaser}
    \vspace{-0.05in}
\end{center}
}]

\begin{abstract}
We propose to perform imitation learning for dexterous manipulation with multi-finger robot hand from human demonstrations, and transfer the policy to the real robot hand. We introduce a novel single-camera teleoperation system to collect the 3D demonstrations efficiently with only an iPad and a computer. One key contribution of our system is that we construct a customized robot hand for each user in the simulator, which is a manipulator resembling the same structure of the operator's hand. It provides an intuitive interface and avoid unstable human-robot hand retargeting for data collection, leading to large-scale and high quality data. Once the data is collected, the customized robot hand trajectories can be converted to different specified robot hands (models that are manufactured) to generate training demonstrations. With imitation learning using our data, we show large improvement over baselines with multiple complex manipulation tasks. Importantly, we show our learned policy is significantly more robust when transferring to the real robot. More videos can be found in the \href{https://yzqin.github.io/dex-teleop-imitation}{project page}.

\blfootnote{
\vspace{-0.05in}
{$\dagger$}Equal advising. Correspondence to: y1qin@ucsd.edu. \\
}
\end{abstract}

\vspace{-1em}
\section{Introduction}
\label{sec:intro}

Dexterous manipulation with multi-finger hand is one of the most challenging and important problems in robotics. The complex contact pattern between the dexterous hand and manipulated objects is difficult to model. It is very challenging to design a controller manually that can solve contact-rich tasks in unstructured environment. Recent research shows possibilities to learn dexterous manipulation skills with Reinforcement Learning (RL)~\cite{Openai2018, Openai2019, Rajeswaran2018}. However, the high Degree-of-Freedom (DoF) joints and discontinuous contact increase the \emph{sample complexity} to train an RL policy. Besides, black-box optimization with RL rewards can also lead to \emph{unexpected and unsafe} behaviors.

Learning from human demonstration collected by teleoperation is a natural solution for dexterous manipulation. Most current teleoperation systems require Virtual Reality (VR)~\cite{Rajeswaran2018, kumar2015mujoco,hedayati2018improving, pan2021augmented, zhang2018deep} devices or wired gloves to capture human hands. While providing accurate data collection, it also limits the flexibility and scalability of the process. On the other hand, vision-based teleoperation frees the human operator from wearing special devices which reduces the cost and is more scalable. 

However, vision-based teleoperation introduces new challenges. It needs to convert the human hand motion into robot hand motion to command the robot, which is called motion retargeting~\cite{handa2020dexpilot, li2019vision, antotsiou2018task}. A human operator needs to choose the next-step movement based on the imagination of the future robot hand gesture and configuration. The human operator may find it hard to control the robot if the retargeting mapping is not intuitive (e.g., controlling a robot hand with less than five fingers), and extra time will be taken to calibrate their own hands with the robot hands. Moreover, the demonstrations collected with a specific robot hand can only be used for imitation learning with the same robot.

In this paper, we introduce a single-camera teleoperation system with a scalable setup and an intuitive control interface that can collect demonstrations for multiple robot hands. Our system only requires an iPad or another mobile device as the capturing device and \emph{DOES NOT need to perform motion retargeting online during teleoperation}. At the beginning, our system will first estimate an operator's hand geometry (\autoref{fig:teaser}, 3nd column in top 3 rows). The key of our system is to generate a customized robot hand on the fly in the physical simulator (\autoref{fig:teaser}, 4th column in top 3 rows). The customized robot hand will resemble the same kinematics structure of the operator's hand in both geometry (e.g., shape and size) and morphology. The system will generate different robot hands for different human operators, providing a more intuitive way for performing dexterous manipulation tasks efficiently.

After the demonstrations are collected with the customized robot hand, we perform motion retargeting via optimization \emph{offline}. We convert the trajectory of a \emph{customized robot hand} to actual \emph{specified robot hands} (i.e., the corresponding models are manufactured and commercialized in the real world). We experiment with 3 types of robot hands including the Schunk Robot Hand~\cite{svh}, the Adroit Robot Hand~\cite{kumar2013fast}, and the Allegro Robot Hand~\cite{allegro} (\autoref{fig:teaser}, last 3 columns in top 3 rows). We only need to collect the trajectories once to generate imitation data for all these specified robots. Note these robot hands can even have different morphology compared to the human hand (e.g., Allegro Hand only has four fingers). We can then use the demonstrations for imitation learning on the corresponding manipulation task. We apply the imitation learning algorithm by augmenting the RL objective with the collected demonstrations in simulation~\cite{Rajeswaran2018}.

We experiment with three types of challenging dexterous manipulation tasks: Relocate, Flip, and Open Door as shown in \autoref{fig:teaser}. By collecting data with our system using the customized robot hand, our user studies show a large advantage over previous methods on efficiency. For example, we can \textbf{efficiently collect around 60 successful demonstrations per hour} for the Relocate task, while directly operating the Allegro Hand in simulation can only collect around 10 successful demonstrations per hour. By imitation learning with the demonstrations collected by our system, we significantly improve dexterous hand manipulation on all specified robot hands over baselines in simulation.

Once the policy is learned in simulation, we can transfer it to the real robot hand. We evaluate with an an Allegro Hand attached on the XArm-6 robot in the real world (\autoref{fig:teaser}, 2 bottom rows). By incorporating human demonstrations into training, our policy learns more human-like natural behavior. Interestingly, this leads to \textbf{much more robust policy when generalizing to the real world and unseen objects}, while pure RL policy fails most of the time.

\section{Related Work}

\begin{figure*}[!t]
    \vspace{-1em}
    \centering
    \includegraphics[width=\linewidth]{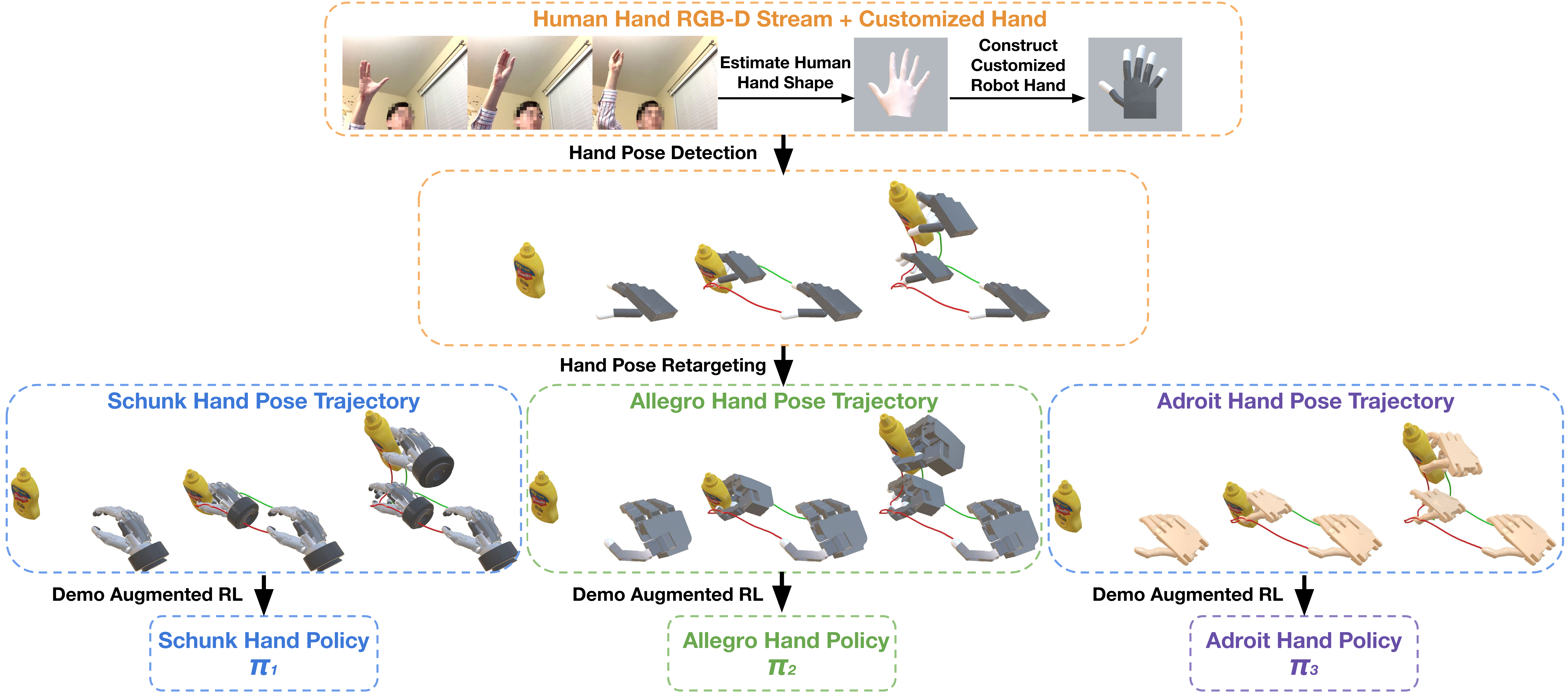}
    \vspace{-0.2in}
    \caption{\small{\textbf{Overall Pipeline:} We stream the hand of human operator with an RGB-D camera. First we construct a customized robot hand in a physical simulator from estimated hand shape parameters result and teleoperate this robot to perform dexterous manipulation task. After teleoperation, we translate the collected trajectory on the customized hand to three different robot hands using retargeting. Finally, we train individual policy on each hand using the translated demonstrations. The red and green curve in 2nd and 3rd rows represent the finger tip trajectory of thumb and pinkie. Different box color means different hand.
    }}
    \vspace{-0.2in}
    \label{fig:main-fig}
\end{figure*}

\textbf{Dexterous Manipulation.} Manipulation with dexterous robot hands has been long studied in robotics and it remains to be one of the most challenging control task~\cite{rus1999hand,okamura2000overview,Andrews2013,Bai2014}. Recently, we have witnessed Reinforcement Learning (RL) approaches~\cite{Openai2018,Openai2019}  delivering promising results on complex in-hand manipulation tasks. While these results are encouraging, RL suffers from poor sample efficiency in training. Under a high degree of freedom (more than 20 in most hands), the RL policy can easily explore unexpected behaviors without well-designed rewards and external constraints. 

\textbf{Imitation Learning from Demonstrations.} Learning from human demonstrations can not only provide external constraint for the robot to explore the expected human-like behaviors but also largely reduces sample efficiency. Beyond behavior cloning~\cite{Pomerleau1989,young2020visual}, imitation learning has been widely studied in the form of Inverse Reinforcement Learning~\cite{Ng2000, Abbeel2004, Ho2016, Fu2017, Torabi2018g, Aytar2018} and incorporating expert demonstrations into the RL objectives~\cite{Peters2008, Duan2016, Vevcerik2017, Rajeswaran2018, radosavovic2020state}. Our work is highly inspired by Rajeswaran \etal~\cite{Rajeswaran2018}, where a VR setup is proposed to collect demonstrations for dexterous manipulation and an algorithm named Demo Augmented Policy Gradient  (DAPG) is introduced for imitation learning. However, data collection with VR requires a lot of human effort and is not scalable. We propose to collect data via a single-camera teleoperation system, which makes the process scalable and accessible for different users. 
Our work is also related to imitation learning from human videos~\cite{schmeckpeper2020reinforcement,shao2020concept,song2020grasping,young2020visual}. However, most of these works focus on a 2-jaw parallel gripper and relatively simple tasks, where 3D information is not necessary. Our teleoperation system provides critical 3D hand-object pose information for guiding dexterous manipulation. 

We emphasize our work is to train a policy that generalizes across different environment configuration, instead of training a policy to follow one expert demonstration, which is proposed in previous motion imitation literature~\cite{Peng2018s, liu2018imitation, Pathak2018, Sharma2018, garcia2020physics, sieb2020graph, xiong2021learning}.

\textbf{Vision-based Teleoperation.}
Vision-based teleoperation frees the operator from wearing data capture devices commonly used in game industry~\cite{zhang2012microsoft} and robot teleoperation~\cite{kofman2007robot, kofman2007robot, du2012markerless, du2010robot, almetwally2013real} on manipulation tasks, e.g. pick and place with a parallel gripper. DexPilot~\cite{handa2020dexpilot} is a pioneering work to extend the vision-based teleoperation to manipulation with an Allegro Hand~\cite{allegro}. To capture the hand pose, a black-clothed table with four calibrated RealSense cameras is used in their system. Our work only requires a single iPad camera for teleoperation. Our novel customized robot hand provides a more intuitive way for data collection and allows generalization for learning with multiple specified robot hands, which has not been shown before.

\textbf{Concurrent Work.} 
One concurrent work from Sivakumar \etal \cite{sivakumar2022robotic} performs vision-based teleoperation to control an Allegro Hand. Their work focus on mapping human hand video to robot control command, and proposed a teleoperation pipeline without using it for imitation learning. In our paper, we provides an end-to-end framework with a more intuitive teleoperation system and a paired imitation learning pipeline for policy training. Another concurrent work from Arunachalam \etal \cite{arunachalam2022dexterous} proposes a teleoperation and visual imitating learning pipeline. They use a 2D hand pose detector during teleoperation, and propose a nearest-neighbor policy to query action from demonstration images. The pipeline proposed in our paper allows human to operate and perform more complex tasks in 3D space, and it can generalize to demonstrations for multiple robot hands.

\section{Overview}
\label{sec:overview}

We propose a novel framework in \autoref{fig:main-fig} to learn dexterous robot hand manipulation from human teleoperation, which is composed by 3 steps as illustrated below.

\textbf{(i) Customized hand teleoperation} is proposed to collect demonstrations for dexterous manipulation tasks. It only requires video streaming from an iPad. A key innovation of the system is constructing a customized robot hand on the fly based on the estimated shape of the operator's own hand. The human operators can then control the customized robot hand in a physical simulation environment to perform dexterous manipulation tasks. The demonstrations can be efficiently collected with around \emph{60 demonstrations per hour}. 

\textbf{(ii) Multi-robots demonstration translation}, which can translated the original demonstration on the customized hand to any dexterous hand that is readily available in the real-world, e.g., Allegro Robot Hand. It computes the state-action trajectory, i.e. joint position and motor command, for the new specified hand that can be consumed by imitation learning algorithm. In our experiments, we try on three dexterous robot hand with different geometry, DoF, and even different number of fingers. 

\textbf{(iii) Demonstration-augmented policy learning} is used to train dexterous manipulation policy on the same task as demonstrations. It augments the Reinforcement Learning objective with behavior cloning using the translated demonstration from (ii). Our framework can efficiently learn dexterous human-like skills on complex tasks which are not well solved by  RL alone. 

We perform Sim2Real transfer on the learned policies with a real Allegro Hand attached on the XArm-6 robot as shown in \autoref{fig:teaser}. In our experiments, we show learning with our demonstrations can significantly increase the robustness of our policy against the Sim2Real gap.

\section{Customized Hand Teleoperation}
\label{sec:teleop_system}

The hardware of our teleoperation system consists of an iPad and a laptop as shown in Figure \ref{fig:setup}. We use the front camera of an iPad to stream the RGB-D video of the human operator at 25 fps. The teleoperation system consists of three components, a physical simulator, a hand detector to capture human motion, and a GUI to visualize the current simulation environment for the human operator. We use a laptop with an RTX 2070 GPU. The processing time for each RGB-D frame is less than 30ms, and the whole teleoperation system can run at 25 fps, the same as camera frequency. 

\subsection{Task Description}
\label{sec:task}
We construct our simulation environment on SAPIEN\cite{xiang2020sapien}, and design multiple dexterous manipulation tasks there. The environments are used for both demonstration collection and policy learning. We develop 3 types of manipulation tasks with different objects. 

\emph{Relocate.} The robot picks up an object and moves it to a target position. It requires the robot to manipulate the object to match the given goal pose. The first row of \autoref{fig:teaser} illustrates the relocate task. We experiment with three objects including \emph{Tomato Soup Can}, \emph{Potted Meat Can} and \emph{Mustard Bottle} from YCB dataset~\cite{calli2015benchmarking}. It is a goal-conditioned task where we \emph{randomize both the initial pose and the goal pose} for each trial. 

\emph{Flip.} As shown in the second row of \autoref{fig:teaser}, it requires the robot to flip a mug on the table. The robot needs to rotate the object 90 degrees carefully to avoid pushing the object away. This task evaluates the robot's ability to exert force towards a certain direction. We \emph{randomize the position and the horizontal rotation along gravity direction} of the mug for each trial. 

\emph{Open Door.} As shown in the third row of \autoref{fig:teaser}. The robot needs to first rotate the handle to unlock the door, and then pull the handle to open the door. The robot needs to grasp the handle with appropriate configuration so that it can achieve both the rotate and pull action. We \emph{randomize the position} of the door for each trial.

\begin{table}[!t]
    \vspace{-0.5em}
	\begin{minipage}{0.46\linewidth}
		\centering
		\includegraphics[width=40mm]{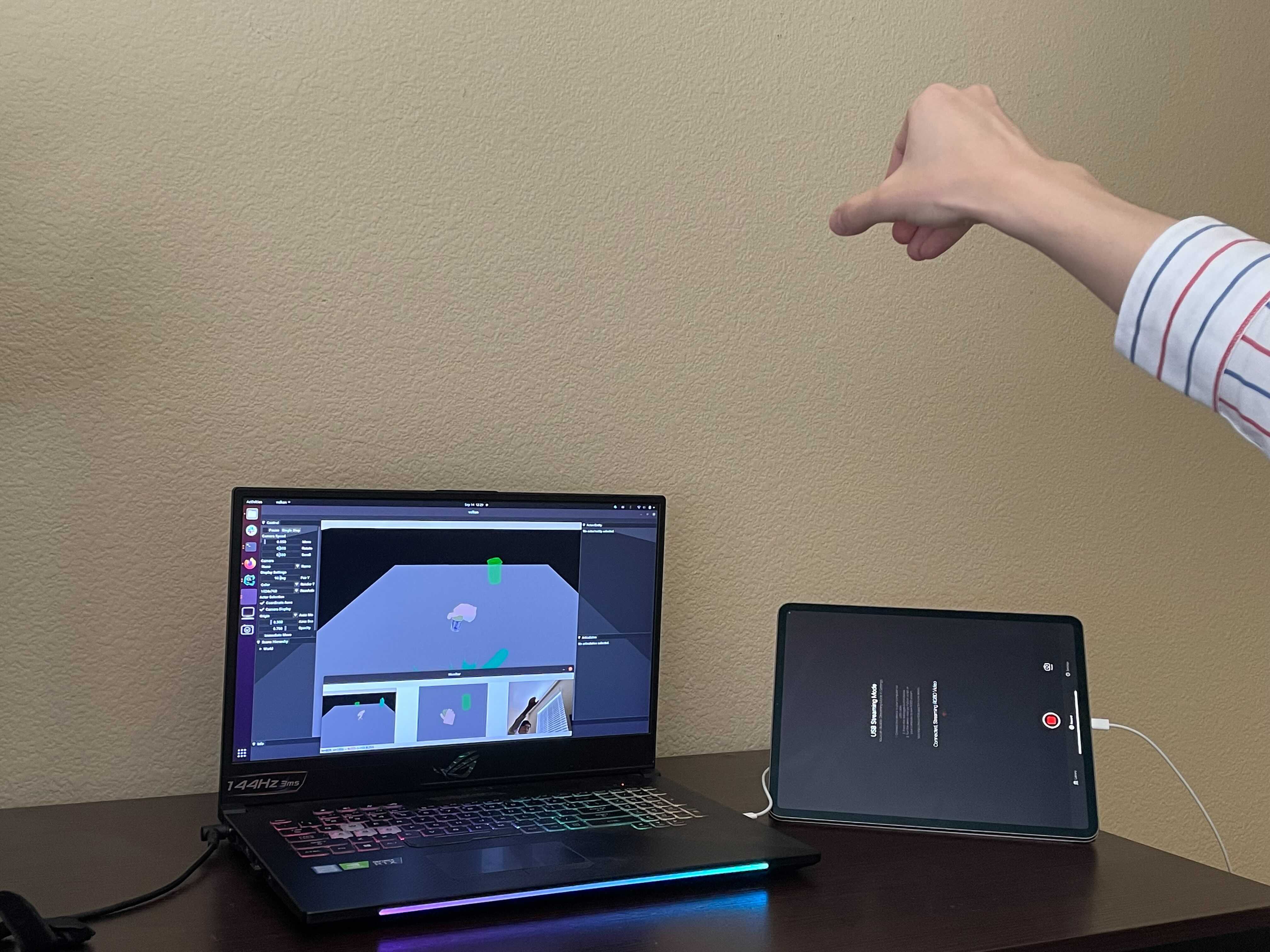}
		\hspace{1em}
		\vspace{-0.15in}
		\captionof{figure}{Hardware setup with an iPad and a computer.}
        \label{fig:setup}
	\end{minipage}
	\hspace{1em}
	\begin{minipage}{0.45\linewidth}
		\centering
		\resizebox{1.1\columnwidth}{!}{%
		\begin{tabular}{l|l}
         Robot & Finger DoF  \\\shline
         Schunk & $ (4,4,3,4,5) $ \\
         Allegro & $(4,4,4,4)$ \\
         Adroit & $ (5,4,4,4,5) $\\
         Customized & $(9,9,9,9,9)$
        \end{tabular}
        }
        \captionof{table}{\small{{DoF comparison for different robot models. Customized stands for the customized robot hand in \autoref{sec:teleop_system}}}}
		\label{tab:dof}
	\end{minipage}\hfill
    \vspace{-0.1in}
\end{table}

\begin{figure}[!t]
    \centering
    \includegraphics[width=\linewidth]{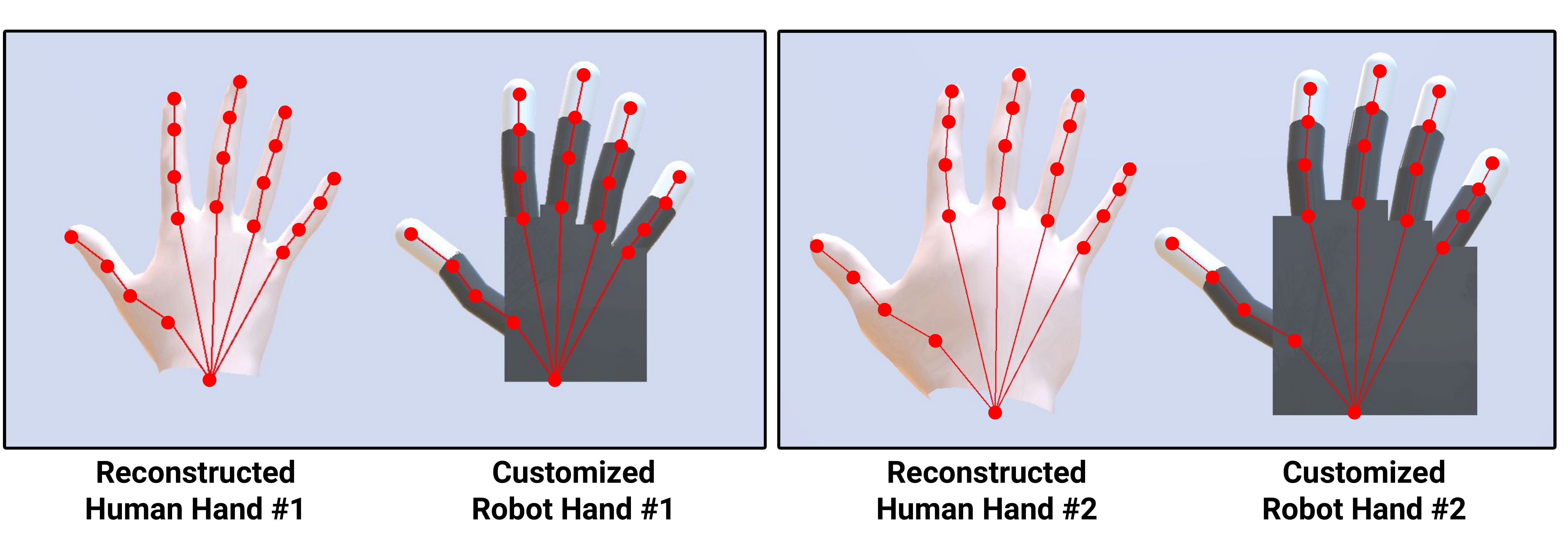}
    \vspace{-0.2in}
    \caption{\small{Illustration of different customized robot hands generated from different human hands. The hand on left and right comes from different human. The red lines visualize the kinematics tree.}}
    \vspace{-0.25in}
    \label{fig:hand}
\end{figure}

\vspace{-0.5em}

\subsection{Hand Detector}
\label{sec:hand_pose}
Our hand detector takes as input the RGB-D frames and outputs the wrist pose, hand pose parameters, and hand shape parameters. It is implemented based upon MediaPipe~\cite{zhang2020mediapipe} and FrankMocap~\cite{rong2020frankmocap}. First, we use MediaPipe hand tracker to detect the axis-aligned bounding-box and crop the image around the hand region. The cropped images are then fed into the pre-trained FrankMocap model to estimate the pose and shape parameters. Frankmocap takes the image as input and regresses the shape and pose parameters of the human hand. It can provide stable results when the hand is not in self-occlusion. We use SMPL-X~\cite{SMPL-X:2019} model to represent pose and shape parameters. It parameters the hand by shape parameters for the hand geometry and pose parameters for the deformation. Given the shape and pose parameters, we can reconstruct a hand in the canonical frame where the wrist is placed at the origin. Then we adopt the Perspective-n-Point (PnP) algorithm to match the key points in the canonical and the detected key points in the camera frame to solve the transformation of the wrist to the camera. The outputs of the hand detector to the downstream modules are wrist pose, hand pose parameters, and hand shape parameters.

\subsection{Customized Robot Hand}
\label{sec:customized}
Our system builds a customized robot hand based on the hand geometry of each user. Given the shape parameters from initialization, we can reconstruct a human hand at rest pose. We then build an articulated hand model in the physical simulator based on the reconstructed human hand. We extract the joint skeleton of the human hand (the red lines in \autoref{fig:hand}) and create a robot model with the same kinematics structure. We choose primitive shapes, e.g. box for the palm and capsules for fingers, for efficient collision detection~\cite{kockara2007collision} and stable simulation~\cite{physx}.  The customized hand has 45 (15*3) DoF, which matches the SMPL-X model. We can rotate the joints of a customized robot hand using detected pose parameters without motion retargeting. \autoref{fig:hand} shows different human hands and the corresponding customized hands. In this figure, the right human hand has a shorter thumb. This characteristic reflects in the customized robot hand.

\begin{figure*}[t]
    \centering
    \includegraphics[width=\linewidth]{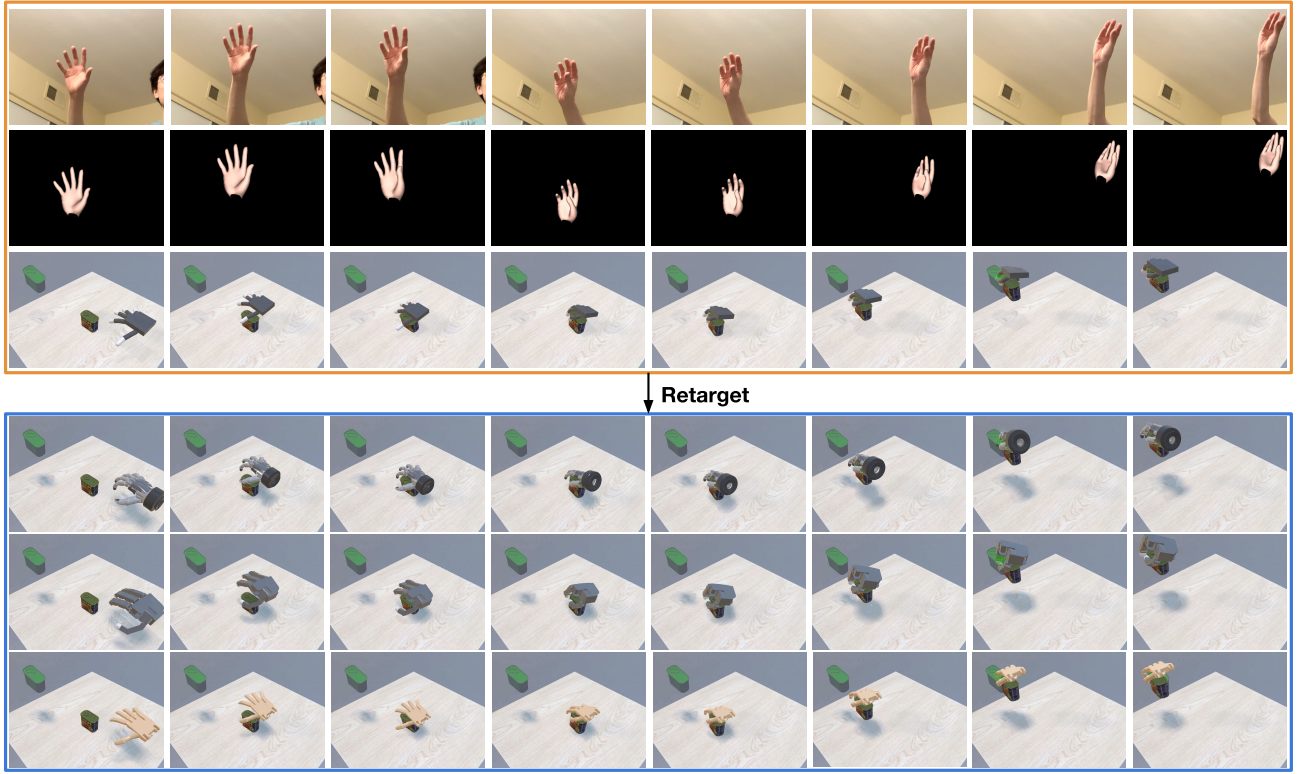}
    \vspace{-0.25in}
    \caption{\small{\textbf{Demonstration Collection and Translation:} The top three rows shows camera stream,  hand pose detection results, and the teleoperated customized robot hand in simulation. The bottom three rows shows the translated demonstration on three different robot hands by retargeting from the teleoperation trajectory.}}
    \label{fig:demo-translate}
    \vspace{-0.2in}
\end{figure*}

We use a PD controller to control the joint angles of the customized robot hand. With each hand detection, we set the estimated pose passed by a low-pass filter as the position target. One challenge of visual teleoperation is the perception error. To tackle this issue, we utilize the hand shape estimation results as a confidence score and use it in PD control. Since the shape parameters are estimated from the best view during initialization, we can use them as the ground-truth hand shape. When the camera view is not reliable, both shape and pose estimation results will suffer from errors. Thus we can compute the error of shape parameters by comparing with the ground-truth and use it as a confidence score for pose accuracy. This problem can be formulated using normal distribution: $s_t \sim \mathcal{N}(s_0, \Sigma)$, where $s_t$ and $s_o$ are the shape parameters from $t$ frame and initialization, respectively. The covariance $\Sigma$ is a diagonal matrix. 
We compute the normalized probability density $p(t)$ as the confidence score. 
The confidence-based PD position control is,
\begin{equation}
    u(t) = p(t) K_p e(t) + k_d \frac{de(t)}{dt},
\end{equation}
where $u(t)$ is the joint torque and $k_p$ and $k_d$ are PD parameters. When the perception error is large, we reduce the stiffness of controller. This design eliminates the undesired abrupt motion caused by perception error.

\section{Multi-Robots Demonstration Translation}
\label{sec:imitation}

\subsection{Hand Pose Retargeting}
\label{sec:retargeting}
Table \ref{tab:dof} shows the DoF of each finger for different robot models. The finger DoF is given in the order from Thumb to Pinky. We need to convert the demonstration from the customized hand to a specified robot, namely hand pose retargeting. With our customized hand, we can skip the computationally-heavy motion retargeting during teleportation and then do it offline. We formulate the motion retargeting as an optimization problem, which is defined based on the keypoints, e.g. tip position,  on the both hand as
\begin{equation}
\begin{split}
\min_{q^R_t} \sum_{i=0}^{N} || f_i^C(q^C_t) - f_i^R(q^R_t)& ||^2 + \alpha||q^R_{t} - q^R_{t-1}||^2\\
\mathrm{s.t.} \quad  q^R_{lower}  \le q^R_t & \le q^R_{upper},
\end{split}
\end{equation}

where $q^C_t$ is joint position at step $t$ for customized robot and $q^R_t$ is the counterpart for the specified robot, e.g. Schunk Robot Hand. We use $f^C_i$ and $f^R_i$ to represent the forward kinematics function of $i-th$ keypoints on the two robots. To improve the temporal consistency, we add a normalization term to penalize the joint position change and initialize $q^C_t$ using the value from $q^C_{t-1}$. After solving the objective above, we can compute the joint position trajectory $q^R_t$ for any specified robot. We apply the hand pose retargeting individually for each specified robot in \autoref{fig:main-fig}.

\subsection{Action Computation}
Hand motion retargeting convert the joint pose trajectory from customized hand to another dexterous hand. To support demo-augmented policy learning, we also need the action for each finger joint. We follow the action estimation procedure in DexMV~\cite{qin2021dexmv} to compute the action, i.e. joint torque or motor control command, from joint pose trajectory of the specified robot. We first pass the joint pose trajectory into a first-order low-pass filter. Then we compute the joint torque via manipulator equation~\cite{murray2017mathematical} of robot inverse dynamics $\tau=f_{inv}(q, q', q'')$. For more details, please refer to \cite{qin2021dexmv}.

\textbf{Visualization on Teleoperation and Translation.} We use \autoref{fig:demo-translate} to summarize the previous two sections. The first three rows show the human hand controlling a customized robot hand in simulator to relocate a can object to a target position (a transparent green shape). Once the demonstration is collected, we can convert it to demonstrations for different robot hands (Schunk, Allegro, Adroit hands) executing the same task in the bottom three rows.

\section{Demonstration-Augmented Policy Learning}

\label{sec:rl}

Given the retargeted demonstration, we perform imitation learning to solve the dexterous tasks defined in \autoref{sec:task}. Naive behavior cloning may be hard to work with randomized initial and target pose. Instead, we adopt imitation learning algorithms that incorporate the demonstration into RL. Specifically, we use Demo Augmented Policy Gradient (DAPG)~\cite{Rajeswaran2018} formulated below as our imitation algorithm.
\begin{align*}%\label{eq:dapg}
    g_{aug}=&\sum_{(s,a)\in\rho_{\pi_\theta}}\nabla \ln\pi(a\vert s) A^{\pi} (s,a)+\nonumber\\
    &\sum_{(s,a)\in\rho_{\pi_\mathrm{demo}}}\nabla \ln \pi_\theta (a\vert s)\lambda_0\lambda_1^k \max_{(s',a')\in\rho_{\pi}} A^{\pi}(s',a'),
\end{align*}
where the first line is the vanilla policy gradient objective in RL and the second line is imitation objective using demonstration. It can be regarded as a combination of behavior cloning and online RL. $\rho_{\pi}$ is the occupancy measure under policy $\pi$, $\lambda_0$ and $\lambda_1$ are hyper-parameters, and $k$ is the training iterations. $A^{\pi}(s',a')$ is the advantage function.

\section{Experiment}
\label{sec:exp}

We first demonstrate the benefits of using the proposed customized robot hand in teleoperation for data collection by a user study with 17 different human operators. Then we show that by leveraging the demonstrations collected by our system, we can improve the policy learning performance by a large margin on various tasks in simulated environment. Finally, we perform real-world experiments on the Allegro hand attached on the X-Arm 6, which shows that the demonstration can improve the policy robustness when transferring to the real-world with higher success rate.

\subsection{Teleoperation User Study}
\label{sec:user_study}

We compare the proposed customized robot hand with the standard robot hand during teleoperation. We ask 17 different human operators to perform \emph{Relocate} and \emph{open door} tasks using 4 different robot hand models: (1) Customized robot hand; (2) Schunk SVH hand; (3) Adroit hand; (4) Allegro hand. For the last three robots, online motion retargeting is required to convert human hand motion onto robot motion while for the customized robot hand we directly use the human pose parameters as the PD target for each joint.

\begin{table}[t]
\begin{tabular}{l|c|c|c|c}
        Robot & \textbf{S.1} success & \textbf{S.1} Time & \textbf{S.2} success & \textbf{S.2} Time \\\shline
        Schunk  & 61.2\% & 14.2s & 30.6\%  & 30.3s  \\
        Adroit  & 58.8\% & 11.5s & 28.2\%  & 37.6s  \\
        Allegro  & 44.7\% & 18.7s & 16.9\%  & 42.5s  \\
        Customized & \textbf{78.9\%} & \textbf{9.1s} & \textbf{55.3\%}  & \textbf{23.0s} 
\end{tabular}
\caption{\small{Success rate and completion time for \emph{Relocate} task. \textbf{S.1} and \textbf{S.2} denotes stage 1 and stage 2.}}

\label{tab:user_relocate}
\end{table}

\begin{table}[t]
\begin{tabular}{l|c|c|c|c}
        Robot & \textbf{S.1} success & \textbf{S.1} Time & \textbf{S.2} success & \textbf{S.2} Time \\\shline
        Schunk  & 83.5\% & 9.2s & 60.0\%  & 20.4s  \\
        Adroit  & 81.2\% & 8.5s & 61.2\%  & 18.9s  \\
        Allegro  & 71.8\% & 12.7s & 41.1\%  & 23.6s  \\
        Customized & \textbf{95.3\%} & \textbf{6.2s} & \textbf{82.4\%}  & \textbf{15.3s} 
\end{tabular}
\caption{\small{Success rate and completion time for \emph{Open Door} task. \textbf{S.1} and \textbf{S.2} denotes stage 1 and stage 2.}}
\vspace{-2em}
\label{tab:user_door}
\end{table}

\begin{figure*}[!t]
    \centering
    \includegraphics[width=\linewidth]{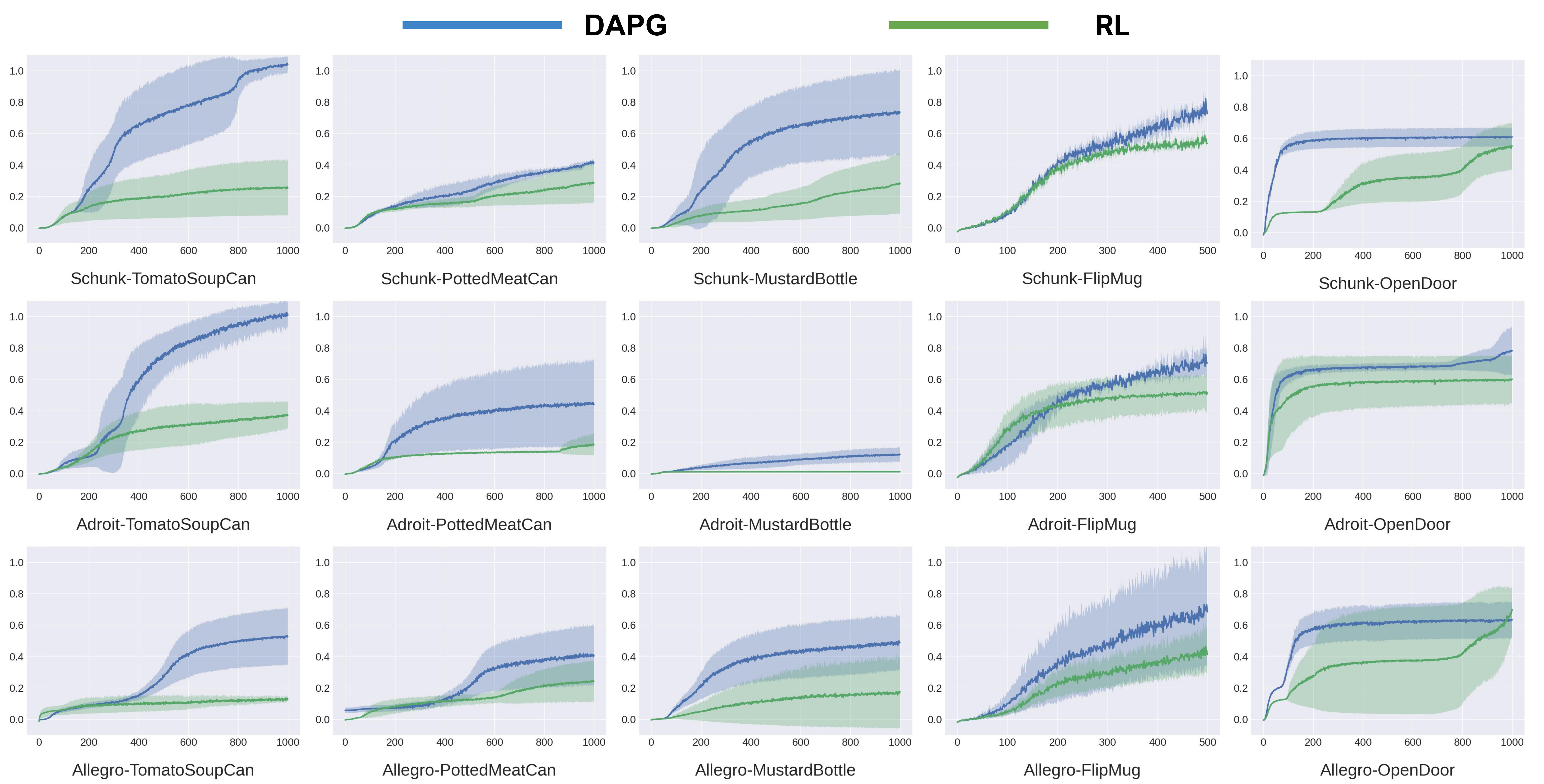}
    \vspace{-0.2in}
    \caption{\small{Learning curves of RL and DAPG on all tasks with four different robot hands in four rows. The first three columns are Relocate task with three objects. Following two columns are Flip Mug and Open Door tasks. The x-axis is training iterations and y-axis is the normalized return. The shaded area indicates the standard deviation for three random seeds.}}
    \label{fig:main-exp}
\end{figure*}
        
\begin{table*}[t]
    \tiny
    \centering
    % \vspace{-0.1in}
    \begin{minipage}[t]{.33\linewidth}
    \tablestyle{2pt}{1.05}
    \begin{tabular}{l|c|c}
        Task & RL & DAPG \\\shline
        Relocate-Toma.  & $45.3 \pm 4.0 $ & $\textbf{85.0} \pm \textbf{12.3}$\\
        Relocate-Pott.  & $6.7 \pm 6.3 $ & $\textbf{41.0} \pm\textbf{ 20.3}$\\
        Relocate-Must.  & $44.0 \pm 20.0$& $\textbf{75.3} \pm \textbf{24.7}$\\
        Flip-Mug & $48.0 \pm 4.3$ & $\textbf{77.3} \pm \textbf{5.0}$  \\
        Open-Door & $55.0 \pm 14.3$ & $\textbf{69.0} \pm \textbf{7.7}$ 
    \end{tabular}
    \caption*{Schunk Robot}
    \end{minipage}
    \hfill
    \begin{minipage}[t]{.33\linewidth}
    \tablestyle{2pt}{1.05}
    \begin{tabular}{l|c|c}
        Task & RL & DAPG \\\shline
        Relocate-Toma. & $41.7 \pm 30.3$ & $\textbf{95.0} \pm \textbf{3.0} $ \\
        Relocate-Pott. & $0 \pm 0$ & $\textbf{53.3} \pm \textbf{37.7}$ \\
        Relocate-Must. & $0 \pm 0$ & $0 \pm 0$ \\
        Flip-Mug & $28.7 \pm 17.7$ & $\textbf{54.7} \pm \textbf{15.3}$ \\
        Open-Door & $58.3 \pm 21.7$ & $\textbf{78.0} \pm \textbf{19.7}$ 
    \end{tabular}
    \caption*{Adroit Robot}
    \end{minipage}
    \hfill
    \begin{minipage}[t]{.33\linewidth}
    \tablestyle{2pt}{1.05}
    \begin{tabular}{l|c|c}
        Task & RL & DAPG \\\shline
        Relocate-Toma. & $0 \pm 0$  & $\textbf{59.7} \pm \textbf{21.3}$ \\
        Relocate-Pott. & $4.3 \pm 3.7$ & $\textbf{38.3} \pm \textbf{21.7}$ \\
        Relocate-Must. & $36.3 \pm 15.3$ & $\textbf{49.7} \pm \textbf{18.3}$ \\
        Flip-Mug & $33.7 \pm 15.0 $ & $\textbf{51.3} \pm \textbf{34.7}$\\
        Open-Door & $\textbf{69.3} \pm \textbf{38.0}$ & $64.7 \pm 14.7$ 
    \end{tabular}
    \caption*{Allegro Robot}
    \end{minipage}
    \vspace{-0.1in}
    \caption{\small{Success rate of the evaluated methods. We use $\pm$ to represent mean and standard deviation over three random seeds. Relocate task has three different objects: tomato soup can, potted meat can, and mustard bottle. The success of \emph{Relocate} is defined based on the distance between object and target. The success of \emph{Flip} is defined based on the orientation of the object. The success rate of \emph{Open Door} is defined based on the joint angle of door hinge.}}
    \label{tab:main-exp}
\vspace{-.2in}
\end{table*}

\textbf{Task Setup.}
Each human operator is asked to perform \emph{Relocate} and \emph{open door} with all four robot hands. Each task-robot pair is tested five consecutive times. For \emph{Relocate} task, the randomly-sampled target position is visualized by a transparent-green shape, as shown on the top-right of \autoref{fig:main-fig}. For each task, the operator will have three-minute trials to get familiar with the task. A common issue is that operators will become more proficient during the testing. They tend to get better results for the task-robot pairs tested later than the former one. For fairness, the order of robot hands to be tested is randomized for each operator.

\textbf{Evaluation Protocols.}
We divide both \emph{Relocate} and \emph{open door} tasks into two stages. For \emph{Relocate}, the first stage is succeeded when the object is lifted up while the second stage is succeeded when the distance between object and target is smaller than a given threshold. For \emph{open door}, the first stage is successful when the door is unlocked by rotating the handle while the second stage is succeed when the door is opened. We will report the average success rate and completion time for each stage of each task. Note that the completion time does not include the time for initialization, which is required for all these four robots to construct the frame alignment between simulated robot hand and real human hand.

\textbf{Results.}
The average success rate and task completion time over all operators are shown in \autoref{tab:user_relocate} for \emph{Relocate} and \autoref{tab:user_door} for \emph{Open Door}. The customized robot hand achieves the highest success rate on all tasks with a large margin compared with the online retargeting method on the other three hands. Considering the initialization process and other overhead, operator can collect around 60 demos per hour for \emph{Relocate} while using allegro hand can only get 10 demos with success. Human operators report that the customized hand is more controllable than other robot hands. One cause is the uncontrollable time consumption required by online motion retargeting. On the laptop with specified in \autoref{sec:teleop_system}, the motion retargeting steps will takes $76\pm 65$ milliseconds. The large variance is caused by iterative optimization in online retargeting. It increases the difficulty of predicting next-step hand motion. By removing the online retargeting using our customized hand, the teleoperation system can provide smoother and more immediate feedback to human operators. We also find the allegro hands perform the worst in most metrics. One possible cause is that allegro hand only has 4 fingers, and it is much larger than other robot hands with $253$ mm length, while the average length is $193$mm for adult males and is $172$mm for adult females.

\subsection{Task Learning Comparison}
\label{sec:exp_comparsion}

We evaluate on the tasks of \emph{Relocate} three different objects, \emph{Flip} a mug, and \emph{Open Door}. We use the processed demonstration to train policy to perform these tasks and compare them with the RL baseline. We ablate how friction, PD Controller Parameters, Object Density, and the number of demonstrations can affect the learning process. For the RL baseline, we use Trust Region Policy Optimization(TRPO)~\cite{schulman2015trust} as the on-policy algorithm. Both policy and value function are $32 \times 32$ 2-layer Multi-Layer Perceptrons (MLPs). The TRPO will use $200$ trajectories for each step. The imitation learning algorithm is DAPG described in \autoref{sec:imitation} with the same hyper-parameters as TRPO. We collect 50 trajectories of demonstration for each task and retarget the motion from customized hand to the specified robot. We train policies with three different random seeds.

The robot state space contains robot joint angles, velocity of hand palm, object position, and orientation at each time step. We include target position for \emph{Relocate} and joint angle of door for \emph{Open Door}. The action space is composed of two parts: hand palm and finger joints. The motion of the palm is controlled by $6$ velocity controllers ($3$ for translation, $3$ for rotation). And the finger joints are actuated by PD position controllers.

\begin{figure*}[t]
% Fig 1
    \begin{minipage}{0.24\textwidth}
    \centering
    \includegraphics[width=\textwidth]{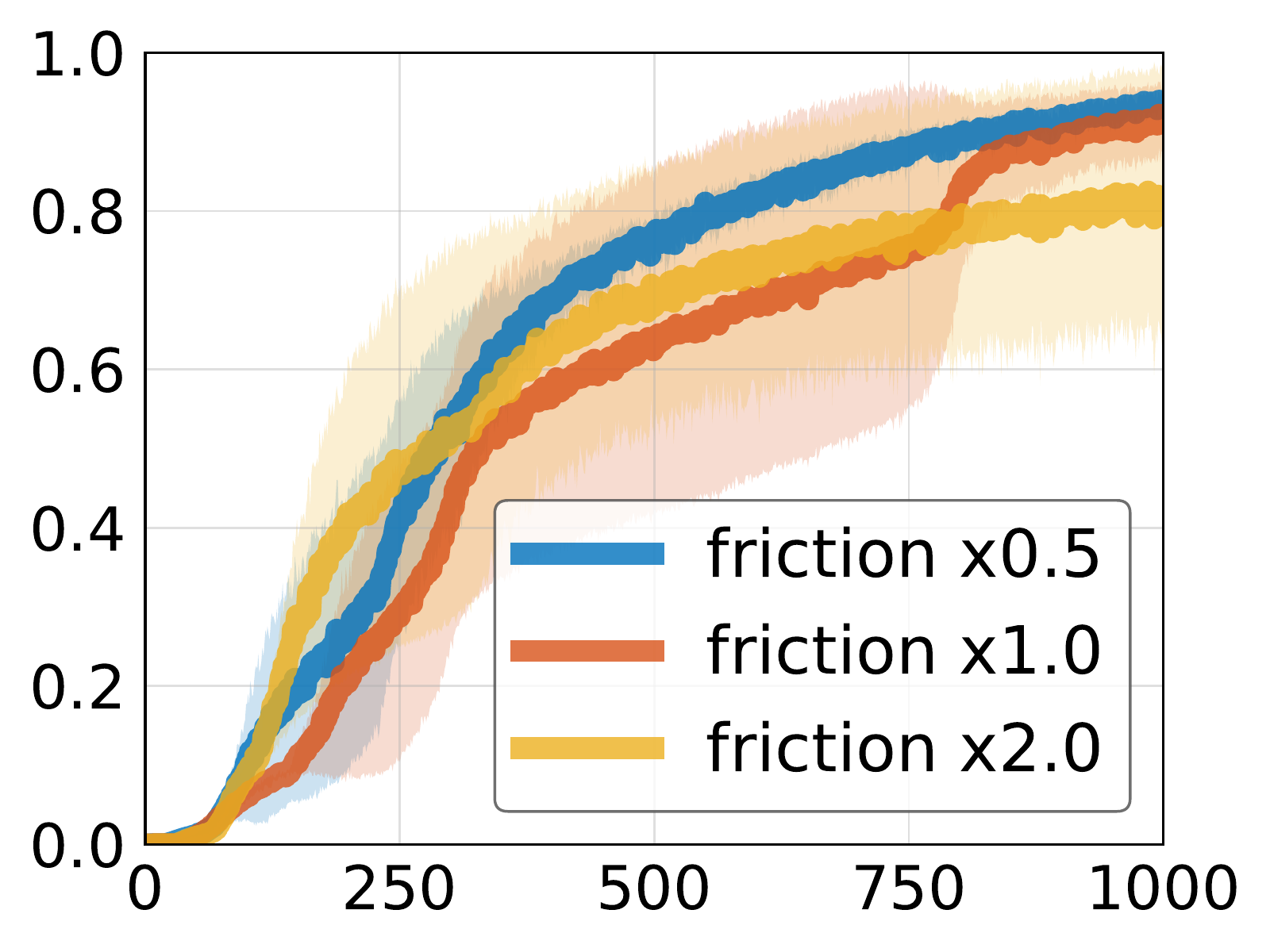} \\
    \quad {(a) Object Friction }
	\end{minipage}
% Fig 3
	\begin{minipage}[h]{0.24\linewidth}
    \centering
    \includegraphics[width=\textwidth]{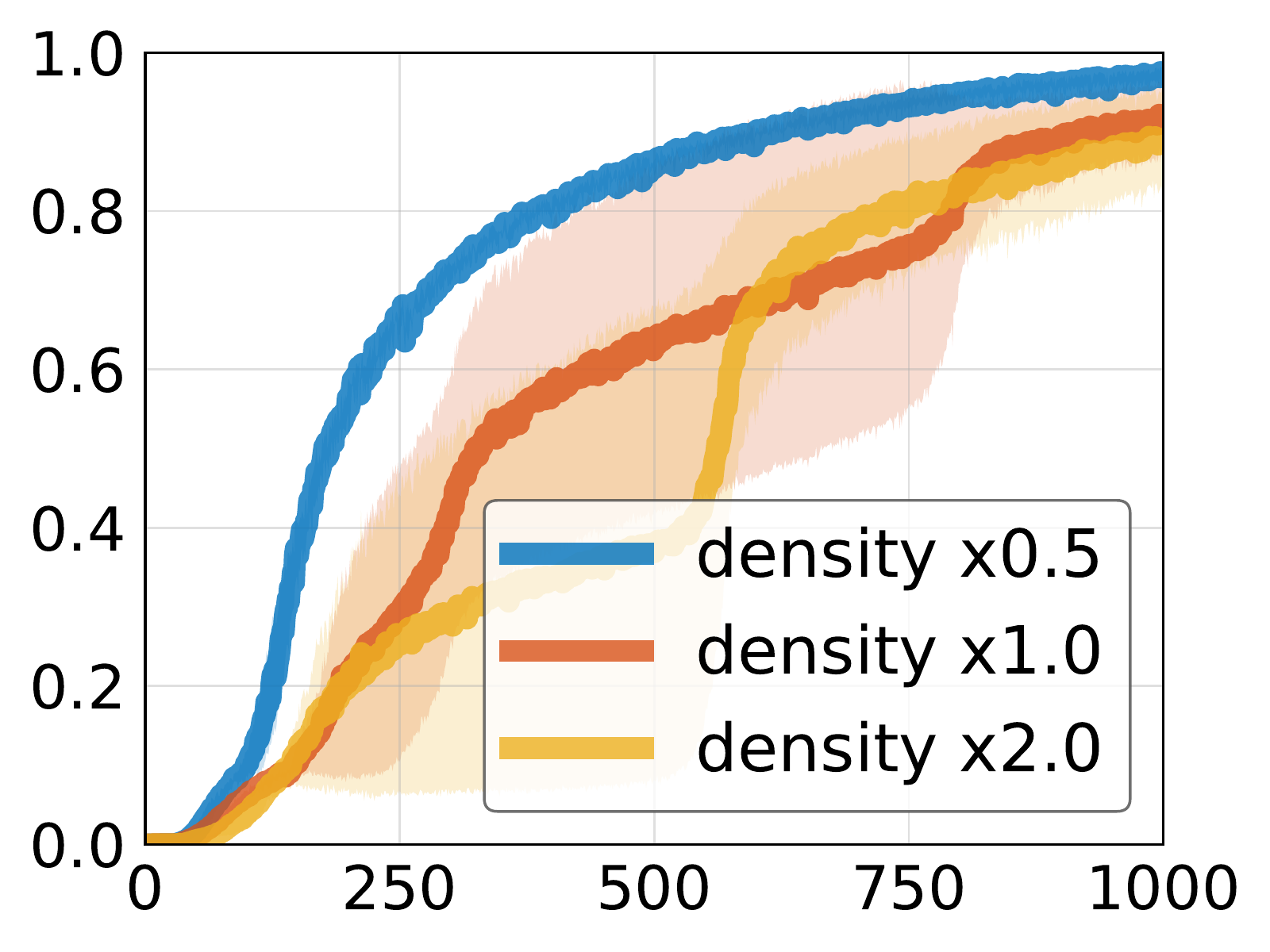} \\
    \quad {(b) Object Density}
    \end{minipage}
% Fig 2
    \begin{minipage}{0.24\textwidth}
    \centering
    \includegraphics[width=\textwidth]{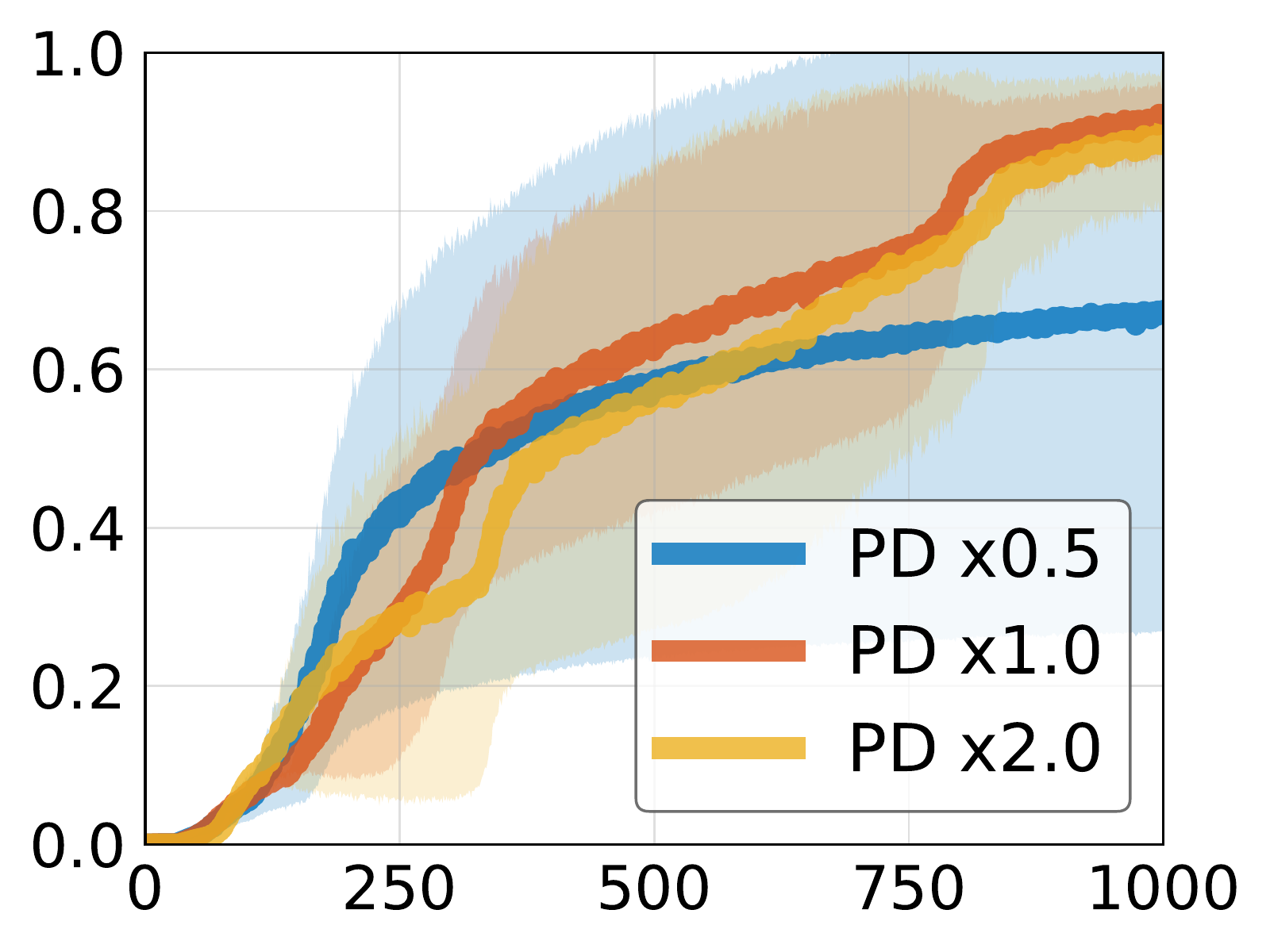} \\
    \quad {(c) PD Controller Params}
	\end{minipage}
% Fig 4
	\begin{minipage}[h]{0.24\linewidth}
    \centering
    \includegraphics[width=\textwidth]{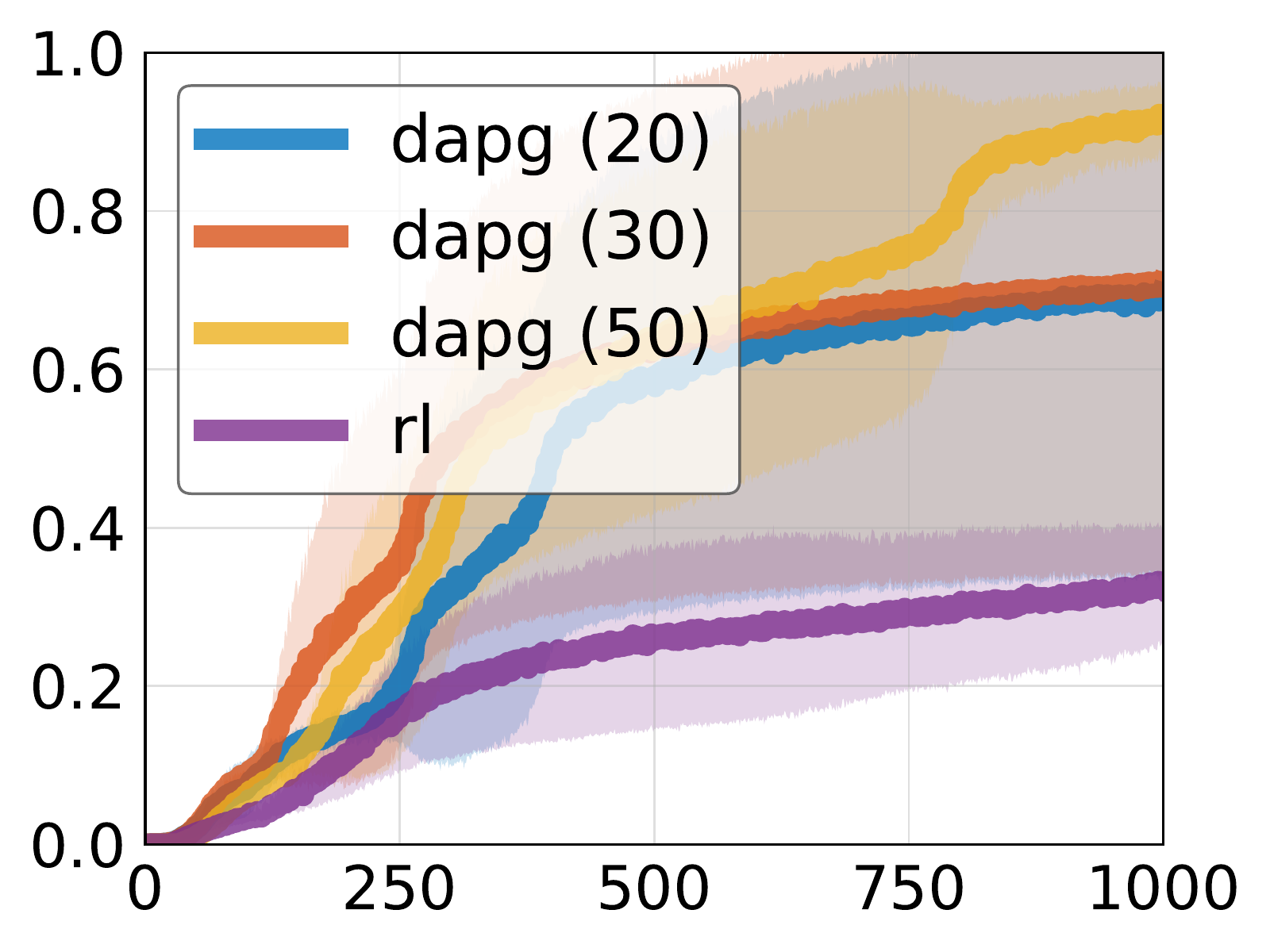}
    \quad  {(d) Num of Demonstrations}
    \end{minipage}%
    \vspace{-0.1em}
    \caption{\small{\textbf{Ablation Study}: Learning curves of DAPG on the \emph{Relocate} task with tomato soup can using Schunk Robot Hand. We ablate: (a) friction parameter of the relocated object; (b) density of object; (c) PD controller parameters: stiffness and damping; (d) number of demonstrations used to train DAPG . The demonstrations are kept the same for all conditions.}}
    \label{fig:ablation}
    \vspace{-0.15in}
\end{figure*}

\begin{figure}[t]
    \centering
    \includegraphics[width=0.8\linewidth]{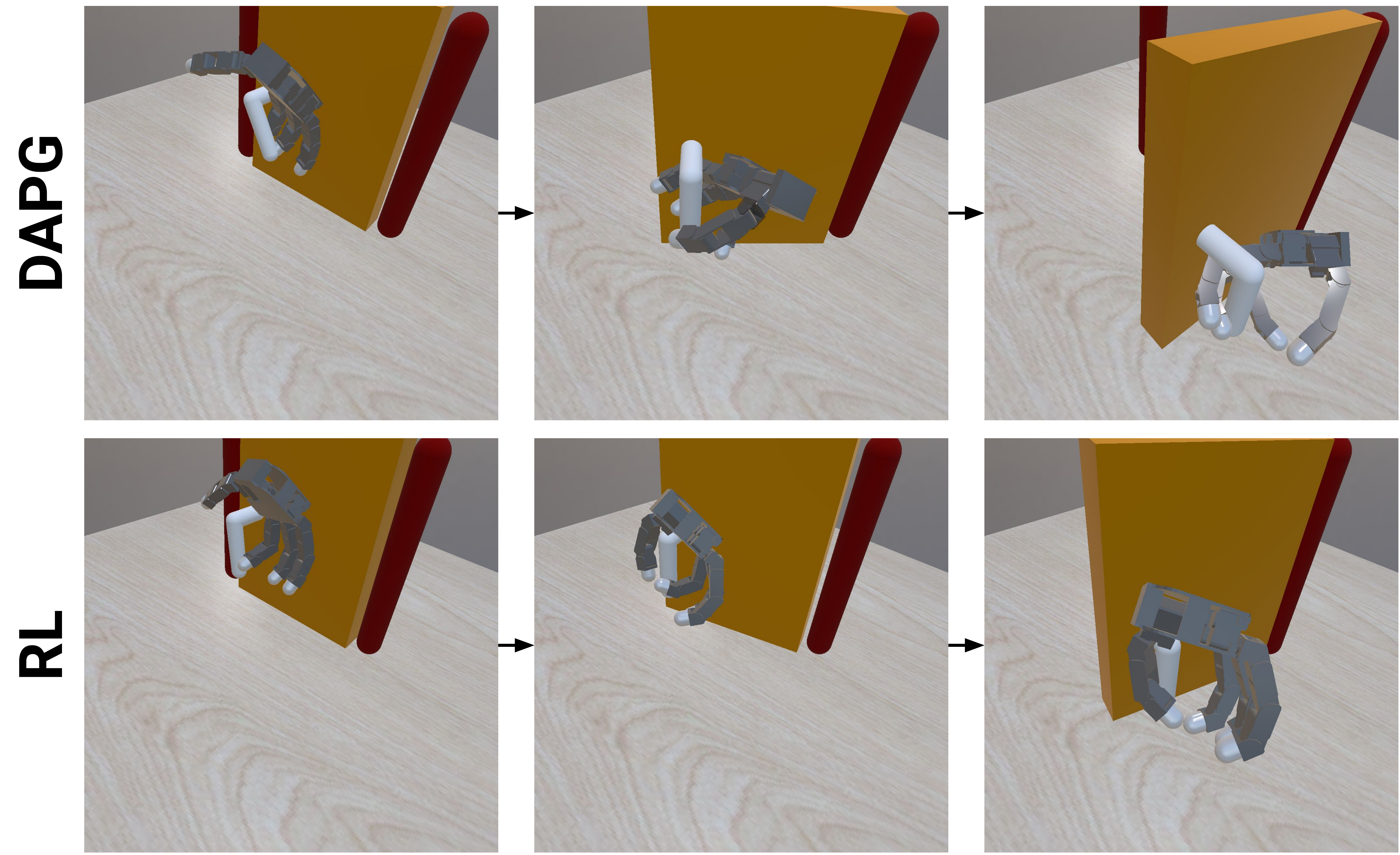}
    \caption{\small{{Comparison of the naturalness on \emph{Open Door} using Allegro Robot Hand. \textbf{Top Row}: policy learned by DAPG with demonstrations; \textbf{Bottom Row}: policy learned by RL without demonstrations.}}}
    \label{fig:natural_policy}
    \vspace{-0.1in}
\end{figure}

We evaluate both RL and DAPG on \emph{Relocate}, \emph{Flip}, and \emph{Open Door} tasks. The training curves are shown in \autoref{fig:main-exp}. The success rate of three specified robot hand is summarized in \autoref{tab:main-exp}. For \emph{Relocate}, the task is considered successful when the object position is within 0.1 unit length to the target at the end of the episode. For \emph{Flip}, the robot will get success when the orientation of mug is flipped back, where the angle between the negative z-axis and the direction of gravity is less than $5$ degree. For \emph{Open Door}, the task is successful when the joint angle of door hinge is larger than $60$ degrees.

As shown in \autoref{fig:main-exp} and \autoref{tab:main-exp}, imitation learning method outperforms the RL baseline for most tasks and robots. For most robots on all tasks, DAPG can outperform pure RL, which shows the demonstration generated by motion retargeting can improve policy training. The only exception is \emph{Open Door} with allegro hand. We visualize the policy trained by DAPG and RL in \autoref{fig:natural_policy}: DAPG tries to open the door by grasping the handle in a natural behavior while RL policy press on the handle with a large force and open the door purely by friction. These results highlight the value of demonstration to regulate the behavior of policy to be expected (human-like) and safe. We find that for both RL and DAPG, \emph{Relocate} with a mustard bottle using Adroit and Allegro robot is very challenging. The reason is that the thumb and other fingers can not form a tight shape closure. While for the Schunk robot, the freedom of the thumb is large enough to grasp the object. On the other hand, Adroit achieves the best on \emph{Relocate} with a tomato soup can. This indicates the existence of robot-specific skills. Different robot hands are designed to fit objects with different geometry and a single robot hand can hardly do best for all tasks.

\subsection{Ablation Study}
To investigate the influence of different dynamics conditions and the number of demonstrations, we ablate the object friction, controller parameters, object density, and the number of demonstrations. We choose the \emph{Relocate} task with tomato soup can using Schunk robot for ablation study. \autoref{fig:ablation} (a) shows that the learning curve is robust to friction change. To hold the object firmly, a two-finger parallel-jaw gripper usually needs to form an antipodal grasp~\cite{chen1993finding}, which is sensitive to friction change. Different from parallel-gripper, the dexterous hand can form force closure with multiple contact points, thus can withstand smaller friction. Similar results can also be found in \autoref{fig:ablation} (b). \autoref{fig:ablation} (c) illustrates the influence of controller parameters. With larger stiffness, the robot can move the object to the target sooner and get a larger reward while controllers with smaller PD can still solve the task. \autoref{fig:ablation} (d) shows more demonstrations can achieve better performance. We also observe that when using only 20 or 30 demos, the variance is larger. 

\begin{figure}[t]
    \centering
    \includegraphics[width=0.8\linewidth]{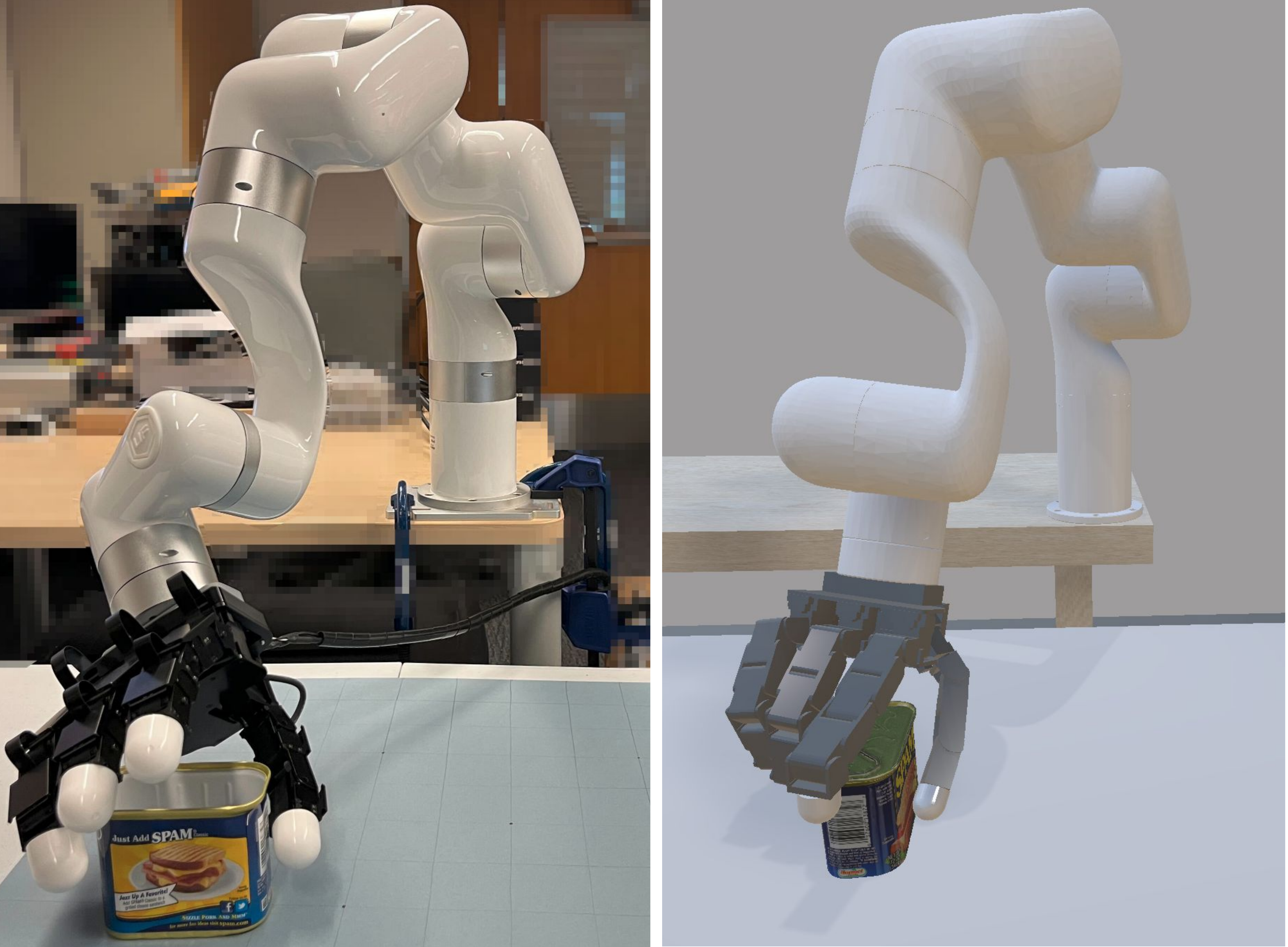}
    \caption{\small{{\textbf{Left side:} real robot setup. The cyan poster on the table is a reference coordinate to determine whether the object is moved to the target position. \textbf{Right side:} simulated robot arm setup.}}}
    \label{fig:real_setup}
    \vspace{-0.1in}
\end{figure}

\begin{figure*}[t]
    \centering
    \includegraphics[width=0.95\linewidth]{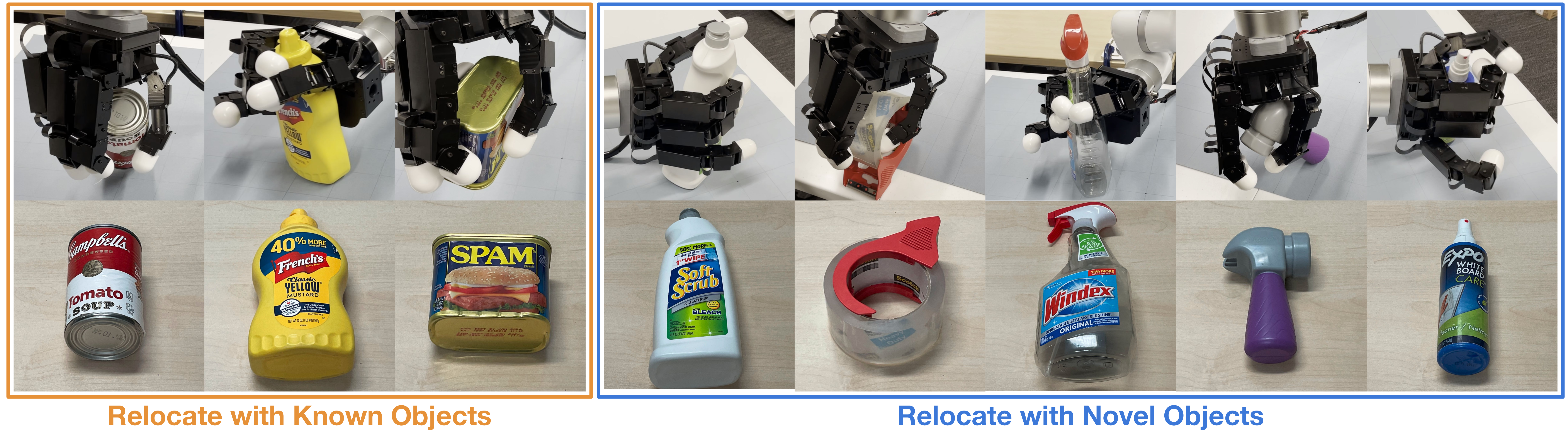}
    \vspace{-0.05in}
    \caption{\small{\textbf{Relocate Task:} Visualization of known objects and novel objects used in our experiments. The first row shows the grasping process and the second row show the object. We test on three known objects and five novel objects that not presented during training.}}
    \label{fig:relocate_novel}
    \vspace{-0.1in}
\end{figure*}

\subsection{Real-World Robot Experiments}

 \begin{figure*}[!t]
    \centering
    \includegraphics[width=0.95\linewidth]{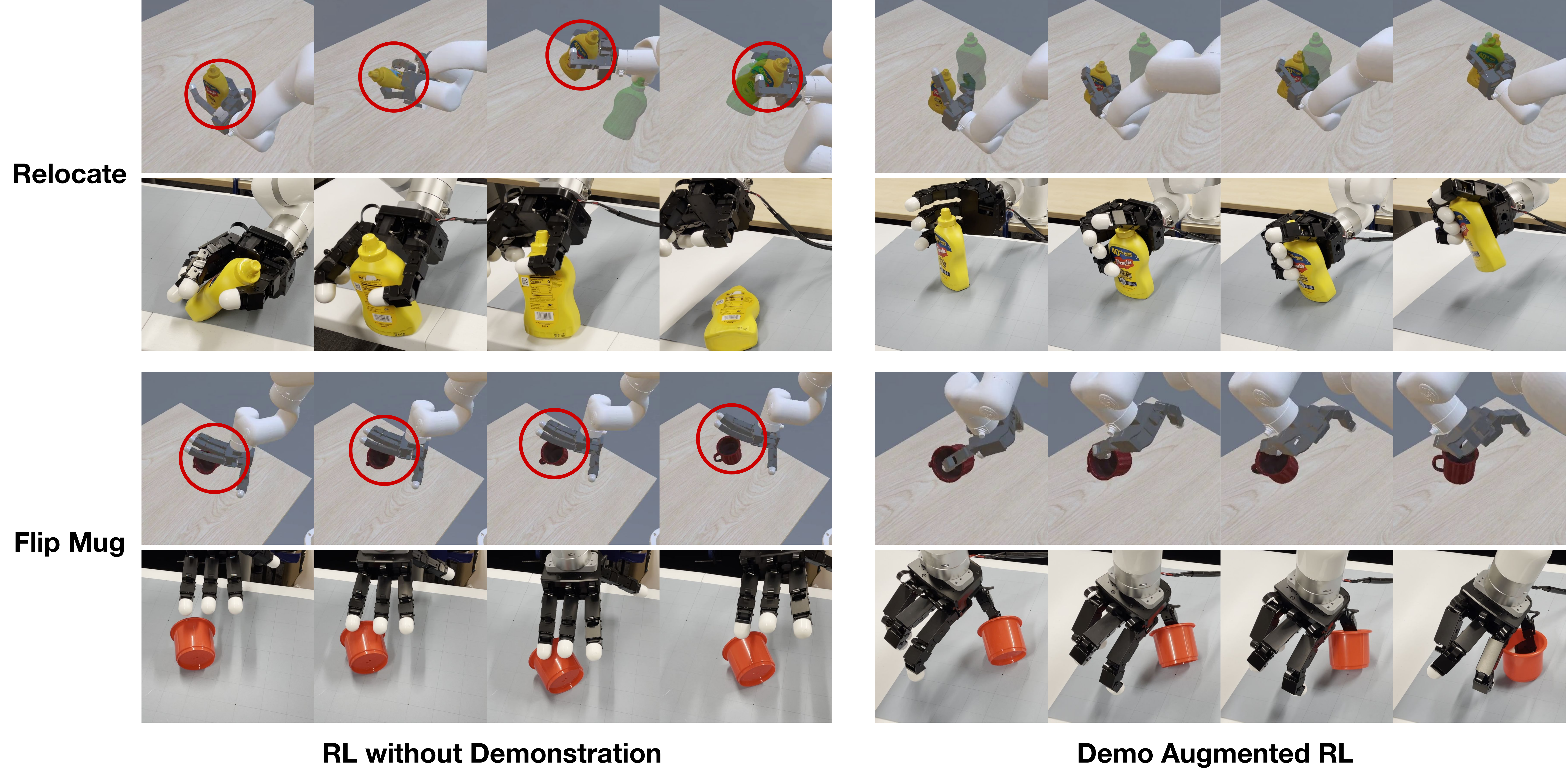}
    \vspace{-0.1in}
    \caption{\small{Policy comparison between pure RL and the demo augmented RL trained with our demonstrations. The RL policy can get success in simulation, but due to its unstable behavior, it can hardly transfer to the real world.}}
    \label{fig:real_comparision}
    \vspace{-0.15in}
\end{figure*}

In the real-world robot experiments, we attached an Allegro hand onto a XArm-6 robot arm~\cite{xarm} instead of using a flying hand. The experiment setup is shown in ~\autoref{fig:real_setup}. We evaluate on the Relocate and Flip tasks. In  simulation, we also build the same XArm6+Allegro model as real-world robot. Similarly as the experiments for flying hand in \autoref{sec:exp_comparsion}, we use the motion retargeting method in \autoref{sec:retargeting} to generate demonstrations and train RL and imitation learning policies in simulation. To facilitate the sim2real transfer, we apply additive Gaussian noise onto the object pose to the observation and randomize the dynamics parameters, such as friction, density, and controller PD parameters during policy training.

\textbf{Task Setup.}
The observation space are composed of robot proprioceptive state, object pose. The target object position is additionally included for Relocate. The object pose in the observation is fixed during the episode and only the initial pose is given, which is estimated by Iterative Closet Point (ICP) algorithm using the point cloud captured by a RealSense D435 camera. 

For \emph{Relocate},  we randomize the initial and target object position for each evaluation trail. The initial object position is randomized within a $20$cm square and the target object position is randomized within a $40$cm square with fixed height. The task is success if the robot can lift the object up to the target position. To determine the final object position, we use the reference coordinate on the table as shown in \autoref{fig:real_setup} to record the xy position of object. If the difference of of xy position between object and target is within 5cm, the trail is considered as a success. In the experiments, we divide the objects into two groups: known object and novel objects. As visualized in \autoref{fig:relocate_novel}, the known object group is composed of three objects that the policies are trained on while the novel object group is composed of five objects that are not seen during training. We evaluate the policy separately on both groups. The task execution sequence is visualized on the top two rows of \autoref{fig:real_comparision}.

For \emph{Flip}, we randomize the initial object position within a $20$cm square. The task is success if the distance between the table top and the highest point on the bottom of mug is smaller than 1cm, which means that the orientation of mug is nearly vertical. The task execution sequence is visualized on the bottom two rows of \autoref{fig:real_comparision}.

\textbf{Quantitative Results.} During evaluation, we randomly sample $9$ object initial and target position pairs and use the same pairs for each policy. For both known object and novel object settings in the Relocate task, we also sampled the object randomly and use the same set of sampled objects for each policy. We evaluate both RL and DAPG policies trained with three different random seeds. The success rate for both tasks is shown in \autoref{tab:relocate_real}. We find when transferred to the real robot, the gap between imitation learning and pure RL is much larger than it is in simulation. We conjecture that a more human-like manipulation policy with imitation learning is more robust to the Sim2Real gap. More interestingly, for the Relocate task, the learned policies can even generalize to novel objects that are not seen in training. Note in our experiments, the geometric shape is not captured by the policy, but only the 6D object pose is. This shows the advantage of multi-finger hand: When operating like human, it offers a certain robustness against the change of shape. 

\begin{table}[t]
    \centering
    \begin{tabular}{l|c|c}
        Task & RL & DAPG \\\shline
        Relocate-Known.  & $22.2 \pm 22.2 $ & $\textbf{77.7} \pm \textbf{11.1}$\\
        Relocate-Novel.  & $18.5 \pm 23.1 $ & $\textbf{66.6} \pm\textbf{11.1}$\\
        Flip Mug & $3.6 \pm 6.4$ & $\textbf{44.4} \pm\textbf{19.2}$
    \end{tabular}
    \caption{\small{Success rate of the evaluated methods on Relocate and Flip tasks. We use $\pm$ to represent mean and standard deviation over three random seeds.}}
    \vspace{-0.2in}
    \label{tab:relocate_real}
\end{table}

\textbf{Policy Visualization in Simulation and Real.} To illustrate why the imitation learning policy is more robust to the physics gap in Sim2Real, we provide visualizations for both tasks in \autoref{fig:real_comparision}. On the top two rows for the Relocate task, we observe that both RL and imitation learning can successfully achieve the task in simulation. However, the RL policy tends to grasp the object using only two fingers with unstable contact (highlighted by red circle). In contrast, policy trained with demonstrations use all four fingers. This leads to a big difference for the real robot: The object slip down from the hand for RL policy while imitation learning policy can stably grasp the object. 

The bottom two rows of \autoref{fig:real_comparision} are the Flip task. Pure RL policy solve the task by pushing on the mug swiftly in simulator while imitation learning policy first place one finger inside the mug and then rotate the wrist. The pushing action from RL policy can hardly success with the real robot hand. While the behavior close to human demonstrations achieve much better results in the real world. 

From both examples, we observe that learning with pure RL for multi-finger hands will try to hack the physics in the simulator and achieve unnatural behaviors, which is hard to transfer to the real world. One the other hand, learning human-like behavior using imitation learning with our demonstrations allows much more robust and stable policy for real world application.

\section{Conclusion}
\label{sec:conclude}

We propose a novel single-camera teleoperation system to collect human hand manipulation data for imitation learning. We introduce a novel customized robot hand, providing a more intuitive way for different human operators to collect data. We show the collected demonstrations can improve the learning of dexterous manipulation on multiple robots and robustness when deployed in real world, when the data collection only needs to be conducted once. 

\balance
{\small
\bibliographystyle{IEEEtran}
\bibliography{./library}
}

\end{document}